\documentclass[letterpaper, 10 pt, conference]{ieeeconf}  

\IEEEoverridecommandlockouts                              

\usepackage{graphicx} 
\usepackage{amsmath} 
\usepackage{amssymb}  
\usepackage{xcolor}

\newcommand{\cstyle}{C_{\text{style}}}
\newcommand{\reals}{\mathbb{R}}
\newcommand{\norm}[1]{\left\lVert#1\right\rVert}

\newcommand{\eref}[1]{Eqn. (\ref{#1})}
\newcommand{\sref}[1]{Sec. \ref{#1}}
\newcommand{\figref}[1]{Fig. \ref{#1}}

\usepackage[sc]{mathpazo}
\usepackage[T1]{fontenc}
\usepackage[font=footnotesize,labelfont=bf]{caption}
\renewcommand{\baselinestretch}{.96}
\usepackage{flushend}

\newcommand{\prg}[1]{\noindent\textbf{#1. }} 

\title{\LARGE \bf
Cost Functions for Robot Motion Style
}

\author{Allan Zhou$^{1}$ and Anca D. Dragan$^{2}$
\thanks{$^{1}$Department of Electrical Engineering and Computer Sciences,
        University of California, Berkeley, Berkeley, U.S.A.
        {\tt\small allan.zhou@berkeley.edu}}%
\thanks{$^{2}$Department of Electrical Engineering and Computer Sciences,
		University of California, Berkeley, Berkeley, U.S.A.}%
}

\begin{document}

\maketitle
\thispagestyle{empty}
\pagestyle{empty}

\begin{abstract}

We focus on autonomously generating robot motion for day to day physical tasks that is expressive of a certain style or emotion. Because we seek generalization across task instances and task types, we propose to capture style via cost functions that the robot can use to augment its nominal task cost and task constraints in a trajectory optimization process. We compare two approaches to representing such cost functions: a weighted linear combination of hand-designed features, and a neural network parameterization operating on raw trajectory input. For each cost type, we learn weights for each style from user feedback. We contrast these approaches to a nominal motion across different tasks and for different styles in a user study, and find that they both perform on par with each other, and significantly outperform the baseline. Each approach has its advantages: featurized costs require learning fewer parameters and can perform better on some styles, but neural network representations do not require expert knowledge to design features and could even learn more complex, nuanced costs than an expert can easily design.

\end{abstract}

\section{Introduction}

Our goal is to enable robots to move more expressively -- communicating internal states like hesitation, or projecting personality or affect aspects like excitement or disappointment. Further, we want robots to do this \emph{while conducting their day-to-day tasks}: we don't want them executing some prescripted motion for communication purposes, only to then go back to the same nominal robotic way when they actually do the task they are supposed to. Rather, style should go hand in hand with the task: opening up the fridge door confidently, moving in towards the juice box cautiously so as to not know over the glass bottle of milk, and happily handing it over to the person. 

Motion style is an active area of research in both robotics, as well as graphics and animation. In graphics, a lot of work has focused on \emph{motion capture style transfer} \cite{torresani2007learning,xia2015realtime,yumer2016spectral,holden2016deep}: taking a clip motion capture data in a certain style and transferring it to another clip. These approaches work well for animated characters, and are really promising for robot motions that happen \emph{outside} of the robot's physical task, like reactions to events. But applying them to achieve style \emph{during} the physical task is challenging, because the robot needs to maintain the constraints that the task imposes.

In robotics, work on expressive motion has focused on the design features that are predictive of style \cite{knight2014expressive,szafir2014communication,sharma2013communicating}, but robots have yet to autonomously generate their motion across tasks instances and task types with style. Autonomous expressive motion generation remains largely confided to expressing intentions, not styles \cite{gielniak2011generating,dragan2013legibility}.

\begin{quote}
\emph{Our observation is that we can capture style through a cost function that augments the robot task objective function and constraints. }
\end{quote}
We explore generating style cost functions for manipulator arms, and leverage trajectory optimization to produce stylized motion using the same cost function across different task instances and types. 

\begin{figure*}
\centering
\includegraphics[width=.7\textwidth]{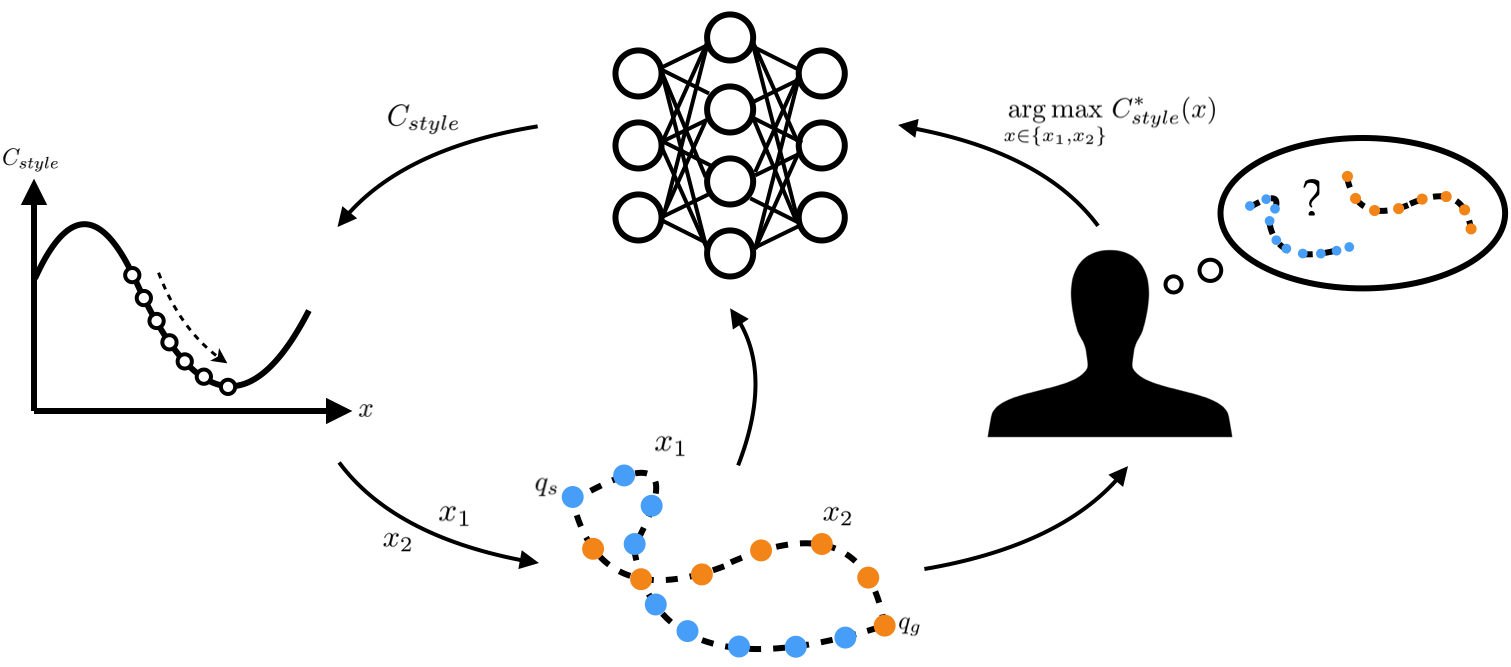}
\caption{
	An overview of our approach to learning style cost functions.
    Counter-clockwise from left: The robot generates trajectories via trajectory
    optimization with the current cost estimate $\cstyle$. The robot
    {\it queries} the expert with pairs of trajectories, and for each pair the expert labels
    the trajectory that is more expressive of the target style. The expert's labels
    are used to update $\cstyle$. We investigate representing $\cstyle$ both as
    a neural network operating on raw trajectory input, and as a weighted linear
    combination of hand-designed features.}
\label{fig:approach}
\end{figure*}

Our work is related to graphics work that learns cost functions for human locomotion styles from demonstration \cite{liu2005learning,lee2010learning}. Unfortunately though, demonstrations of stylized non-anthropomorphic manipulator arms are difficult to acquire,  and features used for learning locomotion style do not transfer to manipulation. 


We make the following contributions:

\prg{Handcrafted style features} We motion capture a dance artist performing day to day manipulation and locomotion tasks in different styles, use observations from this data to design useful features, and learn linear cost functions of these features. 

\prg{Style costs with learned features} While hand-crafted features are useful in that they incorporate domain knowledge, they also rely on a expert to design the right features for them. We thus explore an alternative: learning a cost function represented as a neural network, that operates on the raw trajectory. We contribute an adaptation of deep comparison-based learning to this setting.

\prg{User Study} We compare both featurized and neural network cost functions against nominal motions for different tasks and styles. We find that users rate the cost style-optimized motion as more expressive of the intended style, and that these motions better enable them to identify the intended style. 

The two approaches each have their pros and cons: on the one hand, handcrafting features can be challenging and even though it performed decently on the three styles, we already had difficulty with learning \textit{hesitant} (this is where the neural network performed the best relative to the handcrafted features); on the other hand, neural network representations might have more limited generalization than an expertly designed cost, and tended to slightly underperform for styles that area easier to handcraft features for.

Overall, we are excited to provide an optimization-based approach to autonomously generating stylized motion for robot arms, along with a first attempt at comparing featurized and neural network representations of cost functions for style.

\section{Approach}
In trajectory optimization, we can formulate a motion planning problem
as a constrained optimization problem:
\begin{equation}
\label{eq:trajopt}
\begin{aligned}
& \underset{x}{\text{minimize}}
& & C(x) \\
& \text{subject to}
& & h_i(x) = 0, \; i = 1, \ldots, m.\\
&&& g_j(x) \leq 0, \; j = 1, \ldots, n.
\end{aligned}
\end{equation}
The trajectory $x$ is represented by a sequence of waypoints,
where each waypoint is a particular configuration. If the
robot has $D$ degrees of freedom and the trajectory has $T$ waypoints,
then $x$ is a $D\times T$ matrix where $x[t]$ is a single waypoint.
The constraints $h_i$ and $g_i$ ensure that $x$ completes the task of moving from the
start to goal while avoiding collisions.
$C$ is a cost function that is often designed to encourage minimum-length
paths. One such cost is the sum of squared differences of configurations:
\begin{equation}
    \label{eq:ssd}
    C_{ssd}(x) = \sum\limits_{t=1}^{T-1} \norm{ (x[t+1] - x[t]) }^2
\end{equation}
Problems of the form
\eref{eq:trajopt} can be solved locally by using
trajectory optimizers \cite{schulman2013finding,ratliff2009chomp}.

Given a desired style such as \textit{hesitant}, our goal is to find $\cstyle$
such that doing trajectory optimization on the objective:
\begin{equation}
	\label{eq:overall-objective}
    C(x) = 
    \underbrace{\cstyle(x)}_{\text{Style term}} +
    \lambda \underbrace{C_{ssd}(x)}_{\text{Task term}}
\end{equation}
generates trajectories that complete the task in a hesitant style.
The task term encourages the robot to complete the task
with a reasonably efficient and smooth trajectory, while the style
term encourages the motion to be in the desired style.

\subsection{Featurized Costs}

\begin{figure*}
    \centering
    \includegraphics[width=.8\textwidth]{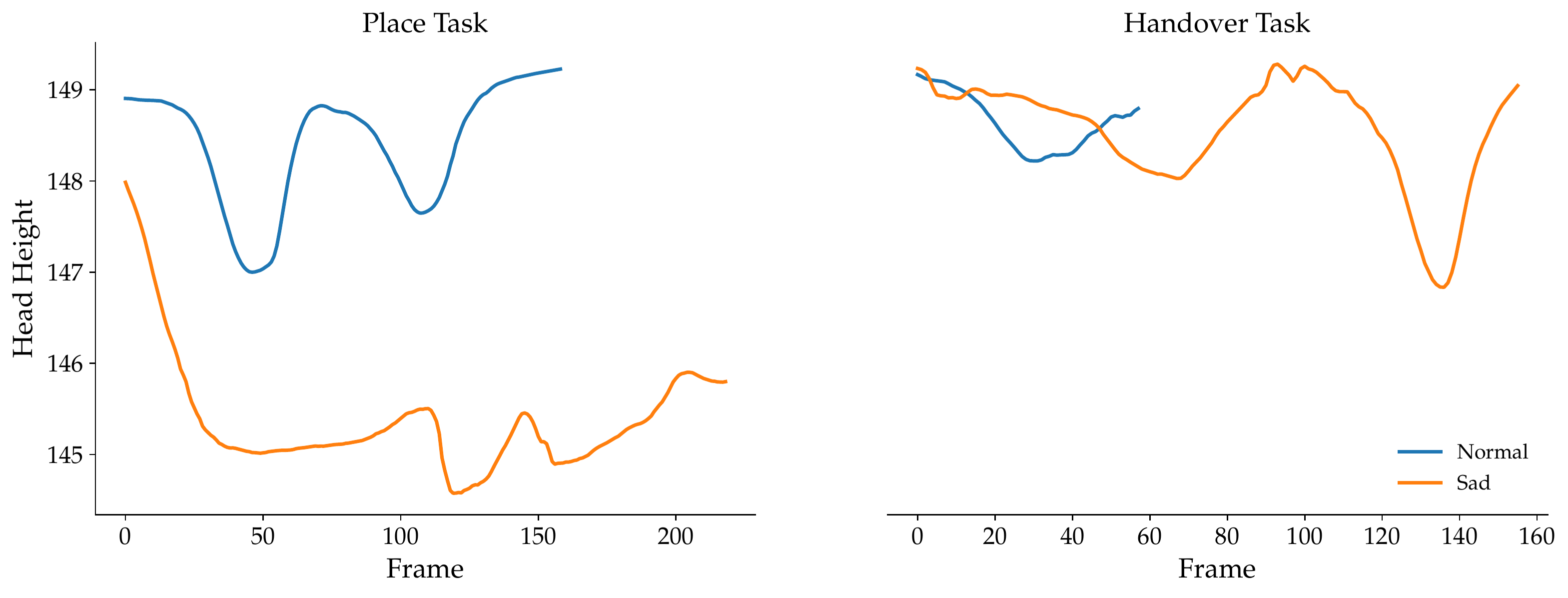}
    \caption{The head height over time of the actress for handover
    and place tasks, as recorded by motion capture.}
    \label{fig:head-height}
\end{figure*}

\label{sec:handcrafted-features}
We first approached the problem by hand-designing trajectory
features that are relevant to a style, and then expressing the
cost as a weighted linear combination of them:
\begin{equation}
    \label{eq:hc-cost}
    \cstyle^{F}(x) = w^T \phi(x)
\end{equation}
where $\phi$ are the features and $w$ are the learned weights.

We identified trajectory features pertaining to three styles:
\textit{happy}, \textit{sad},
and \textit{hesitant}. For \textit{happy} and \textit{sad}, we 
began by studying motion capture data of an actress who performed
different tasks in specified styles. Across different tasks, we
noticed her tendency to ``dip" her head for the \textit{sad} style.
\figref{fig:head-height} illustrates this pattern for both a
handover and place task. As we are working with a non-humanoid
manipulator arm, we naturally decided to focus on features of the
end effector as the robot's analog to a human's head.

Additionally, the actress tended to keep her arms close to the torso
for the \textit{sad} style, while extending them further out for the
\textit{happy} style.
From these observations, we chose to define:
\begin{itemize}
	\item $f_r$: the average horizontal distance from end effector to base (radius)
    \item $f_h$: the average end effector z-coordinate (height)
    \item $f_o$: the average angle between the vertical, or positive z-axis, and the direction the
    end effector is pointing (orientation)
\end{itemize}
We can then express our featurized style costs for \textit{happy} and \textit{sad}
\begin{equation}
\label{eq:happy-sad-cost}
\cstyle^F(x) = \begin{bmatrix}w_r , w_h , w_o\end{bmatrix}^T
\begin{bmatrix}f_r \\ f_h \\ f_o\end{bmatrix}
\end{equation}

For \textit{hesitant}, we included the same three features
$f_r, f_h, f_o$, and added additional features relevant to
timing and motion speed. During
execution the waypoints are equally spaced in time,
so distances between waypoints determine the relative speed
of the robot as it moves. Let
\begin{equation}
	f_{v_i} = \norm{x[i+1] - x[i]}
\end{equation}
A large value of $f_{v_i}$ means the robot will move \textit{quickly}
between waypoints $x[i]$ and $x[i+1]$, while a small value
means the robot will move slowly between those waypoints.
Naturally, we will call our $f_{v_i}$ the \textit{velocity}
features. We then define our featurized \textit{hesitant} cost
in terms of the end effector and velocity features:
\begin{equation}
    \cstyle^F(x) =
    \begin{bmatrix}
    w_r,
    w_h,
    w_o,
    w_{v_1},
    \hdots,
    w_{v_{T-1}}
    \end{bmatrix}^T
    \begin{bmatrix}
    f_r \\
    f_h \\
    f_o \\
    f_{v_1} \\
    \hdots \\
    f_{v_{T-1}}
    \end{bmatrix}
\end{equation} 

\subsection{Neural Network Costs}
\label{sec:nn-costs}
Work in deep reinforcement learning has shown some success in
using neural networks to learn more complicated cost functions
without painstakingly designing complicated features by hand
\cite{wulfmeier2015maximum,finn2016guided,ho2016generative}.
We can parameterize our cost with a Multi-layer Perceptron (MLP) that takes in a
waypoint and forward kinematics information as input, and outputs a vector
$y_t$ for each waypoint of the trajectory. 
We also provide the neural network with velocity and timing information,
using the predecessor waypoint $x[t-1]$ and waypoint number $t$.
Similar to \cite{finn2016guided}, our style cost functions are then of the form
\begin{equation}
    \label{eq:objective-form}
    \cstyle^{NN}(x) = \sum\limits_{t=1}^{T-1}||y_t||^2
\end{equation}
The MLP we chose to parameterize $\cstyle^{NN}$ has two hidden layers with
$tanh$ activations and 42 and 21 units, respectively. The output layer $y_t$
has 21 units is linear (no activation).
We apply Dropout \cite{srivastava2014dropout} after the first two layers during
training.

\subsection{Learning from Preferences}
\subsubsection{Overview}
Both the featurized cost $\cstyle^{F}$ and neural network cost $\cstyle^{NN}$
are parameterized by weights which we
will learn from the human's style cost.
There has been a line of previous work focused on
learning cost functions from humans through demonstrations
\cite{ng2000algorithms,ziebart2008maximum,finn2016guided,ho2016generative}.
Although it may seem natural to obtain demonstrations of styled motion from humans,
demonstrations of styled motion in robots that may be non-humanoid is more difficult.

A useful alternative to learning from demonstrations is learning costs from preferences,
which has been explored for both linear combinations of features \cite{dorsa2017active},
as well as for cases where the cost is parameterized by
neural networks \cite{christiano2017deep}.
In this setup, we repeatedly generate pairs of
trajectories $(x_A, x_B)$, and query the human to pick the
one they think is better. 

In our setting, we are interested in learning a cost for a style,
where task completion is taken care of separately
(by the $C_{ssd}$ term and the trajectory optimization constraints).
If we were learning a cost for, say, the \textit{sad} style, we would
query the human to select the trajectory they think looks more sad.

We assume that the probability the human selects a trajectory
decreases exponentially with cost \cite{bradley1952rank},
so we predict the probability they prefer $x_A$ as:
\begin{equation}
    P(x_A > x_B) =
    \frac{exp(-\cstyle(x_A))}{exp(-\cstyle(x_A)) + exp(-\cstyle(x_B))}
\end{equation}
The human's actual selection serves as the label to the prediction.
Define $I(x)$ as $1$ if $x$ is the human selects $x$, and $0$ otherwise.
The cross entropy loss for this sample is then:
\begin{equation}
    \label{eq:loss}
    L = - I(x_A)\log P(x_A > x_B) - I(x_B)\log P(x_B > x_A)
\end{equation}
For both $\cstyle^F$ and $\cstyle^{NN}$, we can use gradient-based
optimization to update our weights and minimize \eref{eq:loss}.

In our implementation, we generate comparisons in small batches which
are sent to the human for labeling.
Once a batch of comparisons is labeled, we update the weights
using by minimizing the mean loss over the batch.
\figref{fig:approach} shows a visual overview of our approach.

\begin{figure*}
    \centering
    \includegraphics[width=\textwidth]{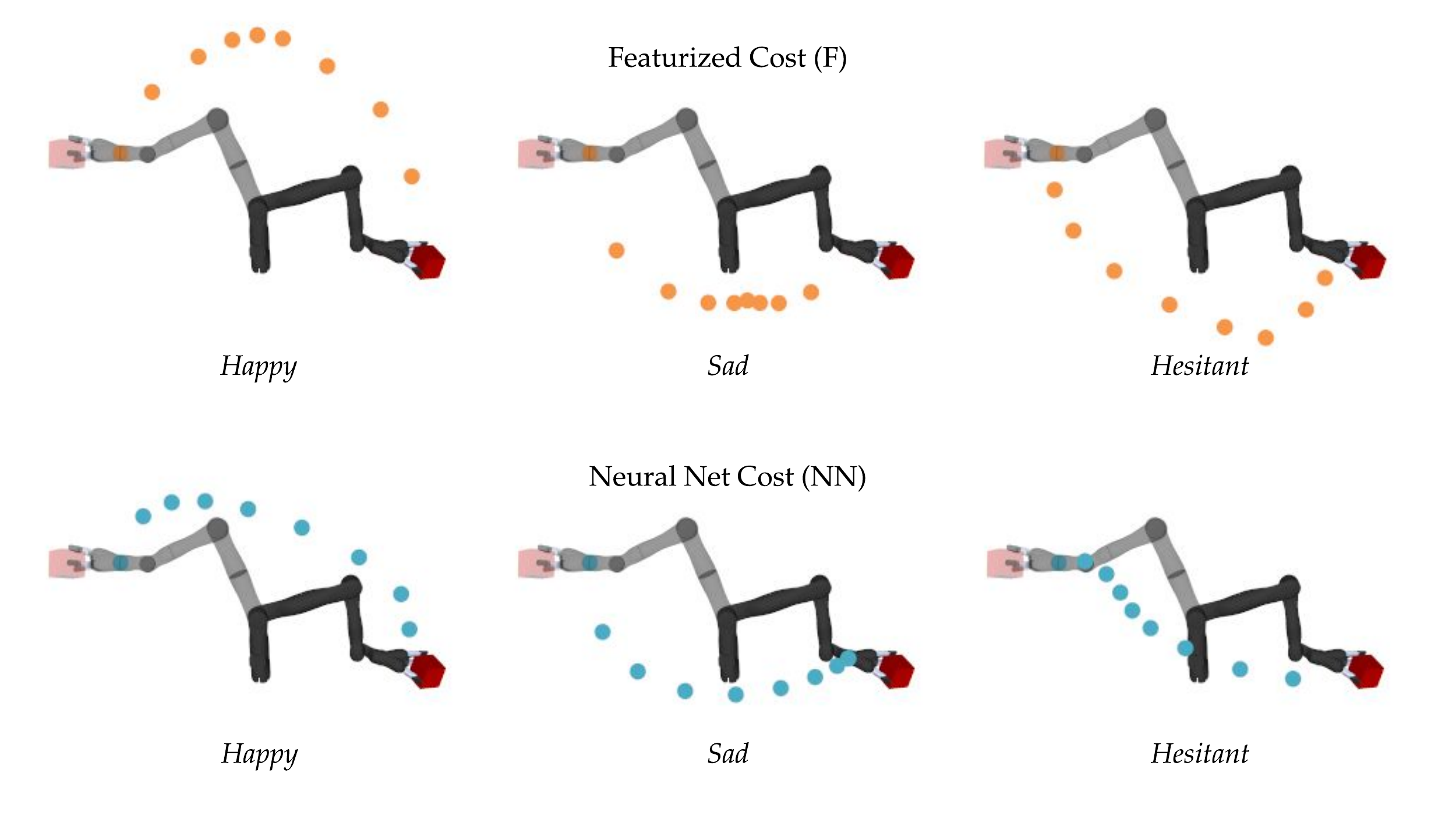}
    \caption{
    Output of optimizing the featurized and neural network costs
    in the place task, for each style.}
    \label{fig:place-f-nn}
\end{figure*}

\subsubsection{Generating Trajectories}
\label{sec:generating}
Given a $\cstyle$, we form an optimization problem
as in \eref{eq:trajopt}, using $\cstyle$ to build the objective according to
\eref{eq:overall-objective}. We use TrajOpt \cite{schulman2013finding} to
find a locally optimal solution $x_0$ to this optimization problem.

Assuming $\cstyle$ is fully learned, we are only interested in the solution
trajectory $x_0$. We can time the trajectory (spacing waypoints equally in time)
and execute it on the robot.

During the learning process, however, we want to produce pairs of trajectories
$(x_A, x_B)$ for the human to compare and label. Moreover, if we only query the
human with trajectories that are optimal with respect to the current $\cstyle$
estimate, the human will have to compare trajectories that all look very similar and
we may not adequately explore the space of possible trajectory styles.
We introduce \textit{exploration} in the learning process by creating additional
trajectories which are random variations of $x_0$. In particular we can create a new
trajectory of the form $x' = x + \Delta$, where $\Delta$ is
a small, smooth perturbation. Repeating this process we create many variations 
of $x_0$, which we use to form query pairs $(x_A, x_B)$.

To create the smooth perturbation $\Delta$, we start with a small random
perturbation $\delta \in \reals^D$ to only a single randomly selected waypoint.
If we think of a trajectory $x$ as a $D\times T$ dimensional vector, with all the waypoints
concatenated together, then:
\begin{gather}
    \Delta_0 = \begin{bmatrix}0 & \cdots & 0 & \delta & 0 & \cdots & 0\end{bmatrix}^T\\
    \label{eq:smoother}
    \Delta = \beta A^{-1}\Delta_0\\
    x' = x_0 + \Delta
\end{gather}
where
\begin{equation}
A = \begin{bmatrix}
2I & -I & 0 & \cdots & 0\\ 
-I & 2I & \ddots & \ddots & \vdots\\ 
0 &  \ddots & \ddots & \ddots & 0\\
\vdots & \ddots & \ddots& 2I & -I\\
0 & \cdots & 0 & -I & 2I
\end{bmatrix}
\end{equation}
The effect of \eref{eq:smoother} is to smooth out the single-waypoint
perturbation $\delta$ to multiple waypoints, and $\beta$ is simply a
scalar coefficient to ensure that the size of $\delta$ does not change.

\begin{figure*}
    \centering
    \includegraphics[width=\textwidth]{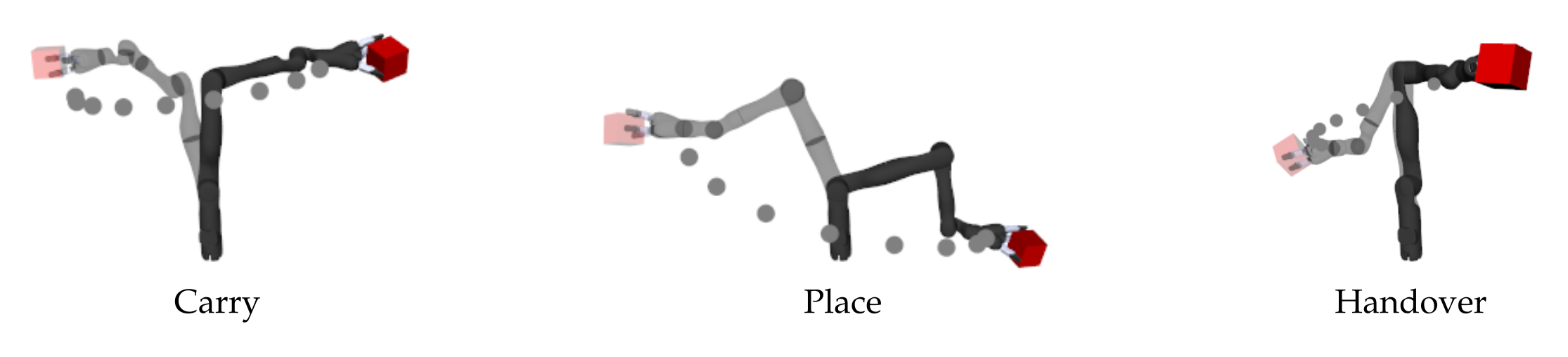}
    \caption{Trajectories generated using sum of squared differences
    cost ($C_{ssd}$) to illustrate each of the 3 tasks.}
    \label{fig:tasks}
\end{figure*}

\subsection{Training Details}
\label{sec:training}
For each of the three styles, we trained a featurized cost $\cstyle^{F}$ 
per \sref{sec:handcrafted-features} and a neural net cost $\cstyle^{NN}$
as described in \sref{sec:nn-costs}. For both cost function types, we
trained with $16$ human labels for the \textit{happy} and \textit{sad} styles
and $15$ labels for the \textit{hesitant} style.

For the neural network cost $\cstyle^{NN}$ in particular, we observed better
performance when we augmented
the data with rotations about the z-axis. Specifically, suppose
the human gives their preference $I(x)$ for a pair of trajectories $(x_A, x_B)$.
Since the robot is upright for all tasks, we assume that the preference does not
change if both trajectories are rotated about the z-axis by the same angle $\theta$.
This means for a single pair of trajectories and the preference label we can
generate multiple training data points with randomly selected $\theta$:
\begin{equation}
    (Rot_z(x_A, \theta), Rot_z(x_B, \theta), I(x))
\end{equation}
We found that this augmentation helped prevent the neural network from
over-fitting to certain start and goal pairs. Note that for $\cstyle^F$,
the features $\phi$ are already invariant to these rotations and so
this augmentation does not have any effect.

\prg{Featurized cost analysis}
For the featurized costs $\cstyle^F$, we can manually inspect the learned weights
on each feature in order to analyze behavior.
For the \textit{happy} and \textit{sad}
styles we learned weights on the
end effector radius, height, and orientation features ($f_r, f_h, f_o$)
as described in \eref{eq:happy-sad-cost}.
The final learned weights for \textit{happy} and \textit{sad} were:
\begin{align}
w^{happy} = \begin{bmatrix}0.03, -0.79, 0.38\end{bmatrix}^T \\
w^{sad} = \begin{bmatrix}0.97,  0.42, -0.50\end{bmatrix}^T
\end{align}
As we expected, $w^{happy}$ puts
a negative weight on the end effector height feature, rewarding
trajectories that move higher. It puts
a positive weight on the orientation feature: since the orientation
feature is the  angle to the vertical, a positive weight penalizes the robot
if its end effector ``dips'' downward. Correspondingly, $w^{sad}$
penalizes end effector height and rewards the
end effector for dipping down.

Meanwhile,
$w^{sad}$ puts a large positive weight on the radius feature,
encouraging the arm to stay closer to the torso and appearing ``withdrawn.''
However, $w^{happy}$ not have the expected corresponding behavior, putting
only a small positive weight on the radius feature.

The learned weights $w^{hesitant}$ act similarly to $w_{sad}$ on the end
effector features.
Interestingly, $w^{hesitant}$ puts negative weights on the velocity
features $f_{v_i}$ for $i <= 4$, and positive weights on $f_{v_i}$ for $i > 4$.
This rewards the robot for
moving more quickly for the first part of the trajectory, and then penalizes
the robot for moving quickly in the second part. The overall effect is that
our featurized cost for \textit{hesitant} learns to encourage trajectories
that slow down closer to the goal.

\section{Experiments}
We run two users studies to evaluate and contrast the featurized
and neural network cost functions trained in \sref{sec:training}
We compared both types of cost functions against the neutral 
non-stylized baseline.

\subsection{Experimental Design}

\subsubsection{Manipulated Factors}

We manipulate the \emph{cost function type} that the robot uses with three levels:
featurized (F), learned neural network (NN), and sum of squared differences (SSD). 

We apply each cost function type to three different styles: \textit{happy}, \textit{sad},
and \textit{hesitant}. For each style, we planned trajectories for
three different tasks:
a carry task, a place task, and a handover task (see Fig. \ref{fig:tasks}).
The task's start and goal configurations were held out during training.
Throughout our experiments we generated trajectories with $10$ waypoints ($T=10)$.
The trajectories generated using each of the learned cost functions in
the place task is shown in \figref{fig:place-f-nn}.

As a simple baseline, we also generated trajectories for each task
using the sum of squared differences cost, $C_{ssd}$.

\begin{figure}
    \centering
    \includegraphics[width=.4\textwidth]{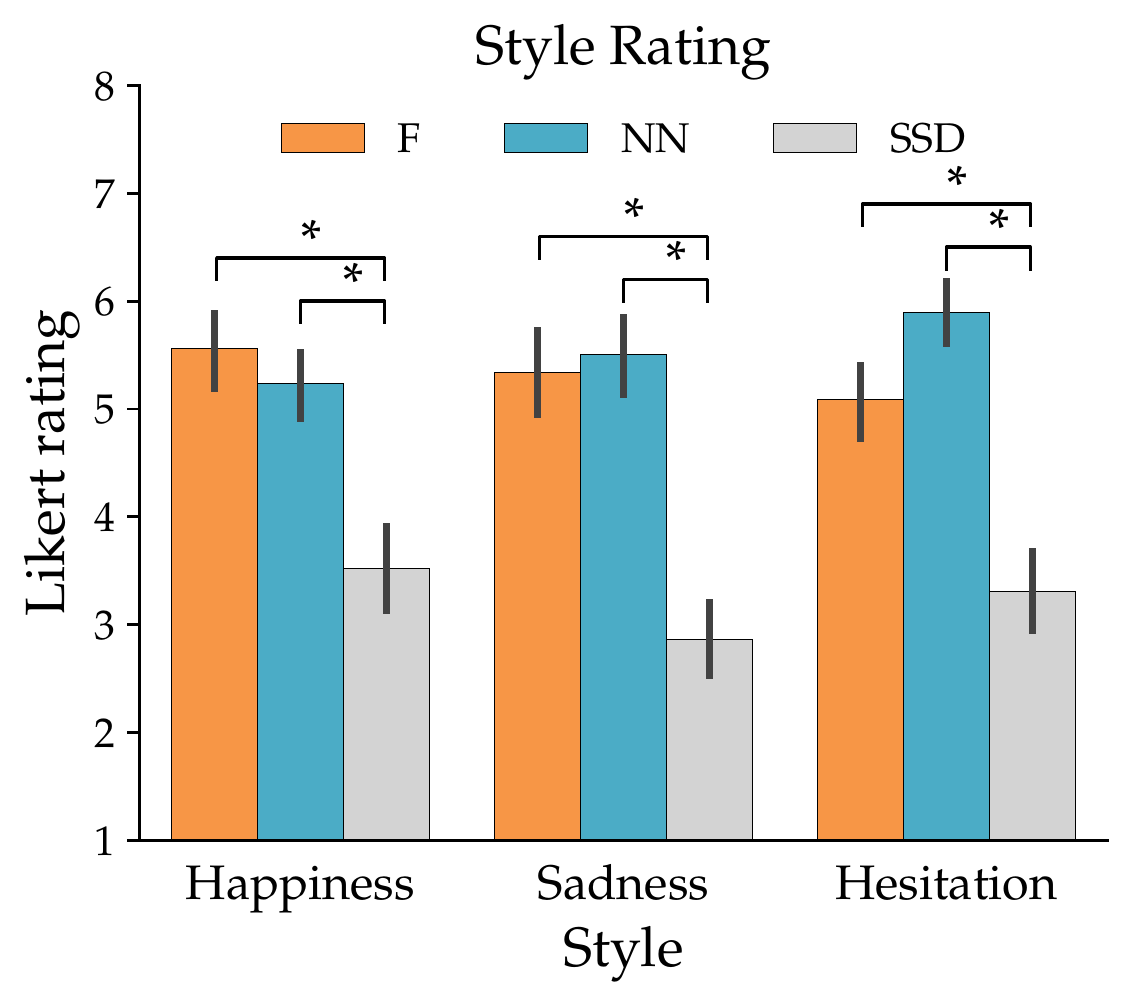}
    \caption{
    Average \textit{style rating} of trajectories produced
    using each type of cost function: featurized (F),
    neural network (NN), and sum of squared differences (SSD).
    The \textit{style rating} is the participant's Likert rating
    on the style the cost function was trained for. For example,
    if a trajectory was generated with a cost trained for \textit{happy},
    the \textit{style rating} is the Likert rating for ``Happiness.''
    }
    \label{fig:exp1-likert-results}
\end{figure}

\subsection{Dependent Measures}
We measure how effective each cost type is at producing motion that has that style. We
measure effectiveness in two ways, in two separate user studies.

\prg{Study 1: Style Rating}
The first study collects ratings of how much the generated motion made the robot
look like the intended style.  Participants saw groups of three motions for
the same task and style, each produced with a different cost type (featurized,
neural network, or SSD). 

First, the participants observed each trajectory and responded to the free
response question ``What style or emotion would you attribute to
this robot?''

We then asked participants to rate on a 7-point Likert scale the happiness,
sadness, and hesitation of each motion they observed, regardless of whether
that trajectory was produced using the corresponding style cost
(to avoid biasing the participants with the question).
For the \textit{style rating} we only took the Likert rating that
matched the cost function. For example, if a trajectory was produced using
a style cost for \textit{happy}, then the \textit{style rating} is the
participant's rating for ``Happiness,'' as opposed to ``Sadness'' or ``Hesitation.''

We also asked a forced-choice question for each style. For instance for 
\textit{happy}, we asked to choose the ``most happy''
motion between three trajectories generated using either the featurized
cost, the neural net cost, or SSD. 

\prg{Study 2: Correct Identification}
The second study tests whether participants can identify the correct style from
distractor styles. In the second study, for each task we presented the
\textit{happy}, \textit{sad}, and \textit{hesitant} trajectories generated using
either the featurized cost or the neural net cost. The trajectory
generated with sum of squared differences was also presented. The
participants were asked to rank each of the four trajectories from
``most \underline{\hspace{.5cm}}'' to ``least \underline{\hspace{.5cm}}''
in each of the three styles.

\subsection{Subject Allocation}
In the first study, both the cost function type and style were
within subjects, while the task was between subjects. Each subject saw
trajectories generated using all cost types
for each of the $3$ styles, but only in a single task.

In the second study, cost function type, style, and task were all within
subjects.

We recruited $60$ ($20$ per task) participants for the first study and
$30$ participants for the second study using Amazon's Mechanical Turk (AMT).
Every participant spoke fluent English and had a minimum $95\%$
approval rating on the AMT platform.

\subsection{Hypothesis}
\emph{We hypothesize that cost type has a significant effect on style rating and correct identification: both of the learned costs should outperform the nominal cost. We do not know which of the two will perform best: the neural network does not have the benefit of hand-designed features, but also has the capacity to learn useful features on its own that we might have not thought about.}

\begin{figure}
    \centering
    \includegraphics[width=.4\textwidth]{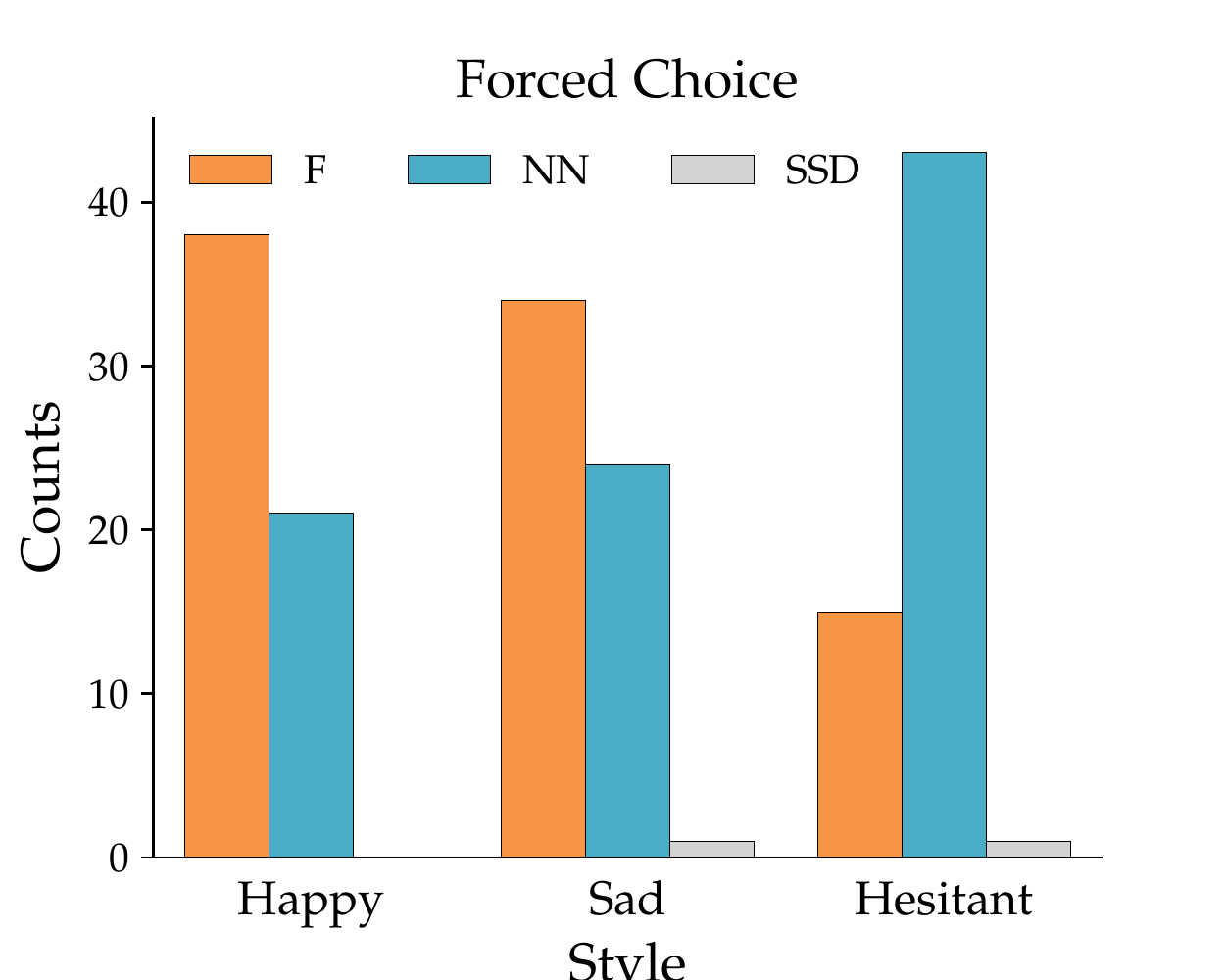}
    \caption{
    Participants compared trajectories in a certain style generated
    using either a featurized (F), neural net (NN), or sum of squared
    differences (SSD) cost function. They then chose the trajectory they thought
    most expressed that style. This plot shows how often trajectories
    generated using each cost were chosen.
    }
    \label{fig:exp1-likert-results-choice}
\end{figure}

\subsection{Analysis}

\begin{figure}
    \centering
    \includegraphics[width=.4\textwidth]{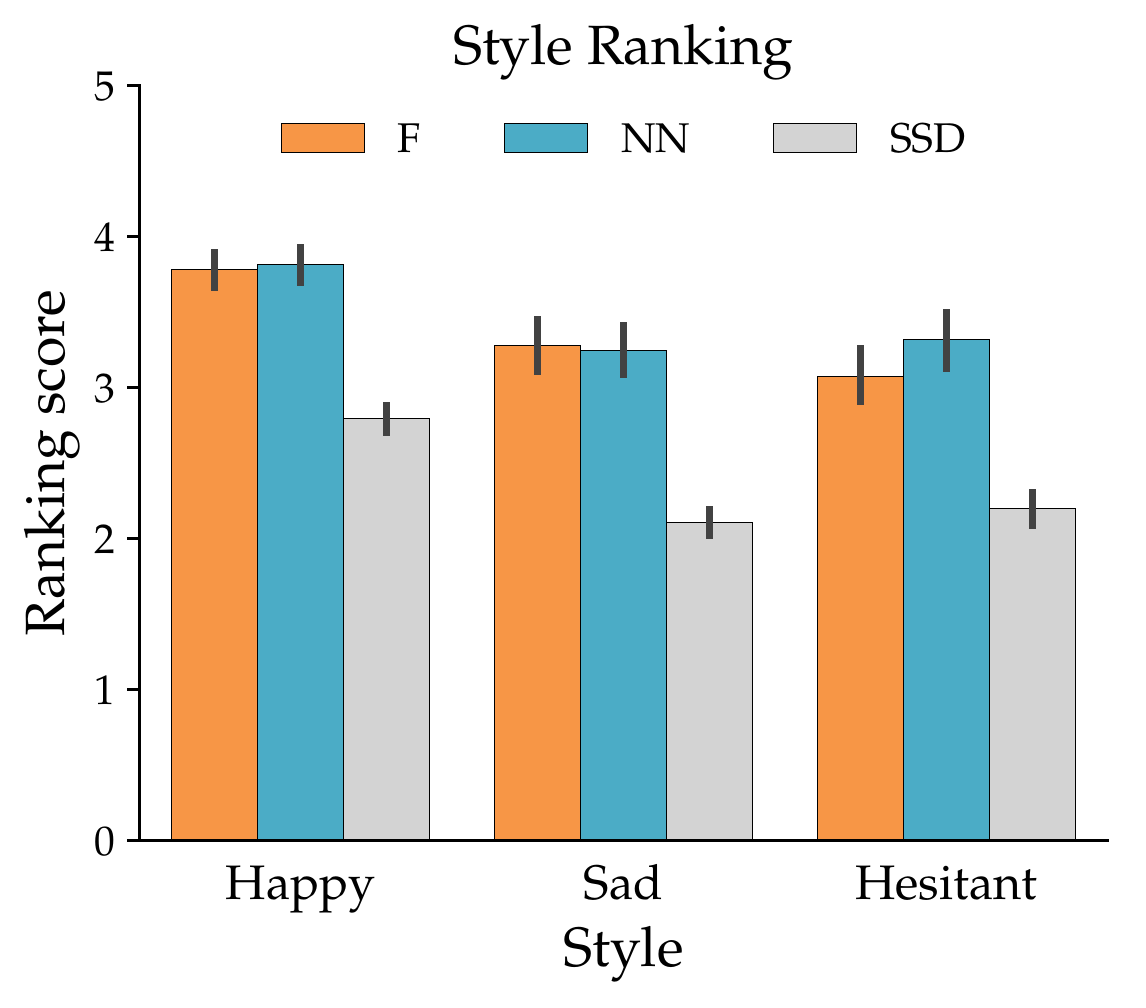}
    \caption{
    The ranking score of each cost function type, which measures
    how successfully participants could correctly identify the style of trajectories
    generated using each cost function type. A higher number means
    more accurate identification.
    }
    \label{fig:exp2-rank-results}
\end{figure}

\begin{figure}
    \centering
    \includegraphics[width=.4\textwidth]{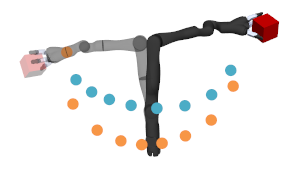}
    \caption{
    The end effector path of \textit{sad} trajectories generated with the
    neural network (blue) cost and the featurized (orange) cost, for the carry task.
    }
    \label{fig:basic-sad-nn-vs-f}
\end{figure}

\begin{figure*}
    \centering
    \includegraphics[width=\textwidth]{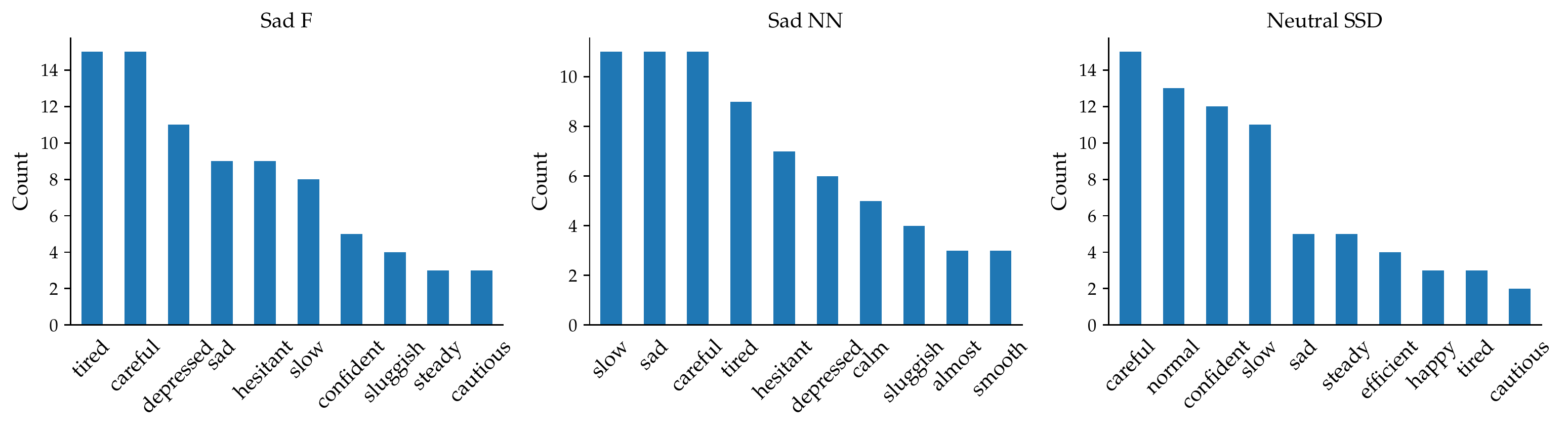}
    \caption{
    Words that participants used to describe robot motion produced
    with the \textit{sad} featurized (F) and neural net (NN) costs,
    as well as the sum of squared differences (SSD) cost.
    }
    \label{fig:sad-word-hist}
\end{figure*}
\prg{Study 1: Style Ratings}
Fig. \ref{fig:exp1-likert-results} plots the results of the \textit{style rating}
from the first study. We analyzed the style ratings using a fully factorial repeated measures ANOVA with cost type, style, and task as factors and user id as a random effect. We found a significant main effect for cost type ($F(2,448)=160.96, p<.0001$), but also interaction effects with the style ($F(4,448)=3.97, p<.01$), task ($F(4,448)=2.45, p=.05$), and the three-way interaction was also significant ($F(8,448)=2.03, p=.04$). We followed up with Tukey HSD post-hocs, and saw that the only significant differences show that both learned costs performed better than the nominal SSD baseline, across tasks and styles. 

There was no conclusive difference between the neural net and the featurized costs. However,
in the forced choice section the featurized cost was preferred for the \textit{happy} and
\textit{sad} styles, while the neural net cost was preferred for \text{hesitant}
(see \figref{fig:exp1-likert-results-choice}). 

\figref{fig:basic-sad-nn-vs-f} plots a comparison of trajectories generated using each
cost type \textit{sad} in the carry task: the
end effector dips much lower in the trajectory generated using the featurized cost,
compared to the one generated using the neural network cost. A corresponding effect is
observed for the \textit{happy} case. Not surprisingly, in simple styles like \textit{happy}
and \textit{sad}, learning weights on a few hand-designed features outperforms the
neural network cost.

The neural net seems to perform best on the hesitant style, judging by the forced choices
results. Indeed, looking at \figref{fig:place-f-nn}, we see that for this style the neural
net cost produces a more sophisticated motion that is slow at first 
and faster after -- this behavior is more nuanced than what the corresponding motion
produced with the featurized cost.

To analyze the responses to the free response questions, we split the responses
into individual words and then removed common ``stop words'' such as ``the,'' ``it,''
``a,'' etc. We also removed the words ``robot,'' ``robots,'' and ``video'' as
participants commonly used them to reference what they were seeing
but they are not relevant to our analysis. After filtering, we plotted the most
commonly used words in a histogram. The histogram for the \textit{sad} style
is shown in \figref{fig:sad-word-hist}. As we would expect from the
\textit{Style Rating} results, the responses to the trajectories generated by
both the neural network and featurized style costs were fairly similar. For both
cases, in addition to ``sad'' and the obvious descriptor ``slow,''
responses commonly referenced styles that are visually similar to
\textit{sad}, such as ``tired'' and ``depressed.''
Meanwhile, the responses
for trajectories generated by $C_{ssd}$ are very different from responses
to the other two types. For example, a common descriptor for these trajectories is
``normal,'' which is not commonly used for the other trajectories.

The pattern for the \textit{happy} and \textit{sad} styles is the same:
responses to trajectories produced with the neural network or featurized costs
are very similar to each other, while the responses to the trajectories
produced by $C_{ssd}$ are different to the other two.

\prg{Study 2: Correct Identification}
We flipped the ranking data so that higher is better, then analyzed it in the same way as the 
style rating. We found a significant main effect for cost type ($F(2,1082)=300.75$, $p<.0001$), with the Tukey HSD posthoc again supporting that SSD has worse performance. Style also had a 
significant main effect ($F(2,1080)=75.83$, $p<.0001$), with the posthoc finding happy to be 
easier to identify. The task was also significant, $(F(2,1088)=3.04, p = .05)$, where
the correct style in the handover task was easier to identify than in the carry task.

There were not significant differences between the neural net and the
featurized costs. These results echo the subjective user ratings.

\section{Discussion}

\prg{Summary}
We approached the problem of generating styled motion in robotic manipulation
tasks by learning two different kinds of styled cost functions. First, we
learned a linear cost function of hand-designed features,
then we learned a neural network cost on raw trajectory input. We trained
both types of cost functions by utilizing human preferences.

We ran two experiments to compare the performance of these two methods, and
the results of the experiments showed that
both methods performed significantly better than neutral trajectories at expressing
the desired style. They also showed some advantages for the neural network cost
in terms of expressing more complicated costs, such as \text{hesitant}, while the
featurized cost matched or beat the neural network cost in simple styles
such as \textit{happy} or \textit{sad}.

\prg{Limitations and Future Work}
This work touches only a small part of the problem of generating styled motion
in robots. We focused on three styles in this paper, but both approaches described
in the paper could be generalized to completely different styles.

As we saw in the experiments, our featurized costs use a rather limited set
of features which could limit their expressivity in some cases. They could be made
more expressive by more carefully considering the style at hand and designing
more complicated features.

Further investigation is also needed to test the preference based learning
system. The styles we investigated required relatively few human responses
to train the neural network cost, but a more difficult style might require
more iterations of the training process and would better test the effectiveness
of the query generation process. The query generation process itself could
potentially be improved to increase the efficiency of the learning process.

\addtolength{\textheight}{-8cm}   




\section*{ACKNOWLEDGMENT}
This work was supported in part by the A.F.O.S.R., the Wagner foundation,
and H.K.U.S.T. We thank the members of the InterACT lab for providing helpful discussion
and feedback.

{
\bibliography{bibliography/IEEEexample}{}

\begin{thebibliography}{10}
\providecommand{\url}[1]{#1}
\csname url@rmstyle\endcsname
\providecommand{\newblock}{\relax}
\providecommand{\bibinfo}[2]{#2}
\providecommand\BIBentrySTDinterwordspacing{\spaceskip=0pt\relax}
\providecommand\BIBentryALTinterwordstretchfactor{4}
\providecommand\BIBentryALTinterwordspacing{\spaceskip=\fontdimen2\font plus
\BIBentryALTinterwordstretchfactor\fontdimen3\font minus
  \fontdimen4\font\relax}
\providecommand\BIBforeignlanguage[2]{{%
\expandafter\ifx\csname l@#1\endcsname\relax
\typeout{** WARNING: IEEEtran.bst: No hyphenation pattern has been}%
\typeout{** loaded for the language `#1'. Using the pattern for}%
\typeout{** the default language instead.}%
\else
\language=\csname l@#1\endcsname
\fi
#2}}

\bibitem{torresani2007learning}
L.~Torresani, P.~Hackney, and C.~Bregler, ``Learning motion style synthesis
  from perceptual observations,'' in \emph{Advances in Neural Information
  Processing Systems}, 2007, pp. 1393--1400.

\bibitem{xia2015realtime}
S.~Xia, C.~Wang, J.~Chai, and J.~Hodgins, ``Realtime style transfer for
  unlabeled heterogeneous human motion,'' \emph{ACM Transactions on Graphics
  (TOG)}, vol.~34, no.~4, p. 119, 2015.

\bibitem{yumer2016spectral}
M.~E. Yumer and N.~J. Mitra, ``Spectral style transfer for human motion between
  independent actions,'' \emph{ACM Transactions on Graphics (TOG)}, vol.~35,
  no.~4, p. 137, 2016.

\bibitem{holden2016deep}
D.~Holden, J.~Saito, and T.~Komura, ``A deep learning framework for character
  motion synthesis and editing,'' \emph{ACM Transactions on Graphics (TOG)},
  vol.~35, no.~4, p. 138, 2016.

\bibitem{knight2014expressive}
H.~Knight and R.~Simmons, ``Expressive motion with x, y and theta: Laban effort
  features for mobile robots,'' in \emph{Robot and Human Interactive
  Communication, 2014 RO-MAN: The 23rd IEEE International Symposium on}.\hskip
  1em plus 0.5em minus 0.4em\relax IEEE, 2014, pp. 267--273.

\bibitem{szafir2014communication}
D.~Szafir, B.~Mutlu, and T.~Fong, ``Communication of intent in assistive free
  flyers,'' in \emph{Proceedings of the 2014 ACM/IEEE international conference
  on Human-robot interaction}.\hskip 1em plus 0.5em minus 0.4em\relax ACM,
  2014, pp. 358--365.

\bibitem{sharma2013communicating}
M.~Sharma, D.~Hildebrandt, G.~Newman, J.~E. Young, and R.~Eskicioglu,
  ``Communicating affect via flight path exploring use of the laban effort
  system for designing affective locomotion paths,'' in \emph{Human-Robot
  Interaction (HRI), 2013 8th ACM/IEEE International Conference on}.\hskip 1em
  plus 0.5em minus 0.4em\relax IEEE, 2013, pp. 293--300.

\bibitem{gielniak2011generating}
M.~J. Gielniak and A.~L. Thomaz, ``Generating anticipation in robot motion,''
  in \emph{RO-MAN, 2011 IEEE}.\hskip 1em plus 0.5em minus 0.4em\relax IEEE,
  2011, pp. 449--454.

\bibitem{dragan2013legibility}
A.~D. Dragan, K.~C. Lee, and S.~S. Srinivasa, ``Legibility and predictability
  of robot motion,'' in \emph{Human-Robot Interaction (HRI), 2013 8th ACM/IEEE
  International Conference on}.\hskip 1em plus 0.5em minus 0.4em\relax IEEE,
  2013, pp. 301--308.

\bibitem{liu2005learning}
C.~K. Liu, A.~Hertzmann, and Z.~Popovi{\'c}, ``Learning physics-based motion
  style with nonlinear inverse optimization,'' \emph{ACM Transactions on
  Graphics (TOG)}, vol.~24, no.~3, pp. 1071--1081, 2005.

\bibitem{lee2010learning}
S.~J. Lee and Z.~Popovi{\'c}, ``Learning behavior styles with inverse
  reinforcement learning,'' in \emph{ACM Transactions on Graphics (TOG)},
  vol.~29, no.~4.\hskip 1em plus 0.5em minus 0.4em\relax ACM, 2010, p. 122.

\bibitem{schulman2013finding}
J.~Schulman, J.~Ho, A.~X. Lee, I.~Awwal, H.~Bradlow, and P.~Abbeel, ``Finding
  locally optimal, collision-free trajectories with sequential convex
  optimization.'' in \emph{Robotics: science and systems}, vol.~9, no.~1, 2013,
  pp. 1--10.

\bibitem{ratliff2009chomp}
N.~Ratliff, M.~Zucker, J.~A. Bagnell, and S.~Srinivasa, ``Chomp: Gradient
  optimization techniques for efficient motion planning,'' in \emph{Robotics
  and Automation, 2009. ICRA'09. IEEE International Conference on}.\hskip 1em
  plus 0.5em minus 0.4em\relax IEEE, 2009, pp. 489--494.

\bibitem{wulfmeier2015maximum}
M.~Wulfmeier, P.~Ondruska, and I.~Posner, ``Maximum entropy deep inverse
  reinforcement learning,'' \emph{arXiv preprint arXiv:1507.04888}, 2015.

\bibitem{finn2016guided}
C.~Finn, S.~Levine, and P.~Abbeel, ``Guided cost learning: Deep inverse optimal
  control via policy optimization,'' in \emph{International Conference on
  Machine Learning}, 2016, pp. 49--58.

\bibitem{ho2016generative}
J.~Ho and S.~Ermon, ``Generative adversarial imitation learning,'' in
  \emph{Advances in Neural Information Processing Systems}, 2016, pp.
  4565--4573.

\bibitem{srivastava2014dropout}
N.~Srivastava, G.~E. Hinton, A.~Krizhevsky, I.~Sutskever, and R.~Salakhutdinov,
  ``Dropout: a simple way to prevent neural networks from overfitting.''
  \emph{Journal of machine learning research}, vol.~15, no.~1, pp. 1929--1958,
  2014.

\bibitem{ng2000algorithms}
A.~Y. Ng, S.~J. Russell, \emph{et~al.}, ``Algorithms for inverse reinforcement
  learning.'' in \emph{Icml}, 2000, pp. 663--670.

\bibitem{ziebart2008maximum}
B.~D. Ziebart, A.~L. Maas, J.~A. Bagnell, and A.~K. Dey, ``Maximum entropy
  inverse reinforcement learning.'' in \emph{AAAI}, vol.~8.\hskip 1em plus
  0.5em minus 0.4em\relax Chicago, IL, USA, 2008, pp. 1433--1438.

\bibitem{dorsa2017active}
D.~Sadigh, A.~D. Dragan, S.~Sastry, and S.~A. Seshia, ``Active preference-based
  learning of reward functions,'' in \emph{Robotics: Science and Systems
  (RSS)}, 2017.

\bibitem{christiano2017deep}
P.~Christiano, J.~Leike, T.~B. Brown, M.~Martic, S.~Legg, and D.~Amodei, ``Deep
  reinforcement learning from human preferences,'' \emph{arXiv preprint
  arXiv:1706.03741}, 2017.

\bibitem{bradley1952rank}
R.~A. Bradley and M.~E. Terry, ``Rank analysis of incomplete block designs: I.
  the method of paired comparisons,'' \emph{Biometrika}, vol.~39, no. 3/4, pp.
  324--345, 1952.

\end{thebibliography}
\bibliographystyle{bibliography/IEEEtran}}


\end{document}